\documentclass[final,5p,times,twocolumn]{elsarticle}
\usepackage{graphicx} 
\usepackage{balance}
\usepackage{cuted, ragged2e}
\usepackage{float}
\usepackage{amsmath} 
\usepackage{wrapfig}

\usepackage{graphicx}
\usepackage{amssymb} 
\usepackage{lineno}
\usepackage{algorithm2e} 
\usepackage{amsmath}
\usepackage{siunitx}
\usepackage{tabularx}
\usepackage{algorithm2e}
\usepackage{amsmath,algpseudocode}
\usepackage{algcompatible} 
\usepackage{graphicx}
\usepackage{lscape}
\usepackage{verbatim}
\usepackage{color,soul}
\usepackage{xcolor}
\usepackage{subfigure}
\usepackage{tabularx,ragged2e,booktabs,caption,array,multirow,multicol}
\usepackage{hyperref}
\usepackage{csquotes}
\usepackage{mathtools}  
\usepackage{listings}
\usepackage{amssymb}
\usepackage{latexsym}
\usepackage{epsfig}
\usepackage{float}
\usepackage{xspace}
\usepackage{algorithmwh}
\usepackage{float}

\newcommand{\mb}[1]{\mathbf{#1}}
 \newcommand{\bs}[1]{\boldsymbol{#1}}

\graphicspath{ {Images/} }
\title{ 
    Langevin-gradient parallel tempering  for Bayesian  neural learning}
 
\author{Rohitash Chandra , Konark Jain , Ratneel V. Deo , Sally Cripps}   

\begin{document}

\thispagestyle{empty}
\pagestyle{empty}

\begin{abstract}
Bayesian  neural learning feature  a rigorous approach to estimation and uncertainty quantification via the posterior distribution of weights that represent knowledge of the neural  network. This not only provides point estimates of optimal set of weights but also the ability to  quantify uncertainty in decision making using the posterior distribution.  Markov chain Monte Carlo (MCMC) techniques are typically used to obtain sample-based estimates of the posterior  distribution. However, these techniques face challenges in convergence and scalability, particularly in settings with large datasets and network architectures. This paper address these challenges in two ways. First, parallel tempering is used used  to explore multiple modes of the posterior distribution and   implemented in  multi-core computing architecture. Second, we make within-chain sampling schemes more efficient by using Langevin gradient information in forming Metropolis-Hastings proposal distributions. We demonstrate the techniques using time series prediction and pattern classification applications. The results show that the method not only improves  the computational time, but provides better prediction or decision making capabilities  when compared to related methods. 
\end{abstract}
\maketitle

\section{Introduction}
 
 Although  backpropagation  neural networks have gained immense attention and success for a wide range of problems \cite{rumelhart1988learning},  they face a number of challenges such as   finding suitable values of hyper-parameters  \cite{zeiler2012adadelta,cawley2007preventing,bengio2000gradient} and appropriate network topology \cite{lawrence1998size,white1993gannet}.   These challenges remain when it comes to different neural network architectures, in particular  deep neural networks architectures which have a large number of parameters \cite{schmidhuber2015deep}. Another limitation of current techniques is  the lack of uncertainty quantification in decision making or prediction.  Bayesian neural networks  can address most of these shortfalls. Bayesian methods
account for the uncertainty in prediction and decision making via the posterior distribution. Note that the posterior is the conditional probability that determined  after taking into account the prior distribution and the relevant evidence or data    via sampling methods.  Bayesian methods can account for the uncertainty in parameters (weights) and topology by marginalization over them in the predictive posterior distribution, \cite{mackay1992practical, mackay1996hyperparameters}.   In other words, as opposed to conventional neural networks, Bayesian neural learning use probability distributions to represent the weights \cite{mackay1995probable,robert2014machine}, rather than single-point estimates by gradient-based learning methods.  
  Markov Chain Monte Carlo   methods (MCMC)   implement Bayesian inference that sample  from a probability distribution \cite{hastings1970monte,metropolis1953equation} where a Markov chain is constructed  after a number of steps such that the desired distribution  becomes the equilibrium distribution \cite{raftery1996,van2016simple}. In other words, MCMC  methods provide 
 numerical approximations of multi-dimensional integrals  \cite{banerjee2014hierarchical}.  Examples of MCMC methods include the Laplace 
 approximation \cite{mackay1992practical}, Hamiltonian Monte Carlo 
 \cite{neal2012bayesian}, expectation propagation 
 \cite{jylanki2014expectation} and variational inference   \cite{hinton1993keeping}.  

Despite the advantages, Bayesian neural networks face a number of  challenges  that include efficient proposal distributions for convergence, scalability and computational efficiency for larger network architectures and datasets. Hence,  number of attempts have been proposed  for enhancing Bayesian neural learning with optimization strategies to form proposals that feature  gradient information  \cite{li2016preconditioned,liang2007annealing,kocadaugli2014nonlinear,Chandra2017Langevin}.  Moreover, in the case of deep learning where thousands to millions of weights are involved, approximate Bayesian learning methods have emerged. Srivastava \emph{et  al.} \cite{srivastava2014dropout} presented "dropouts" for deep learning where the key idea was  to randomly drop neurons   along with their connections  during training  to prevent  over-fitting. Gal  \emph{et  al.} used the concept of  dropouts in a Bayesian framework  for uncertainty quantification in model parameters that feature weights and network topology for deep learning \cite{gal2016dropout} which was further extended for computer vision problems  \cite{kendall2017uncertainties}. Although promising for uncertainty quantification, it could be argued that the approach does not fully approximate  MCMC based sampling methods typically used for  Bayesian inference. 

 Parallel tempering  \cite{marinari1992simulated,geyer1995annealing}     is a  MCMC method that  features    multiple replica Markov chains that provide  global and local exploration  which makes them suitable for irregular and multi-modal distributions \cite{patriksson2008temperature,hukushima1996exchange}. Parallel tempering carries out an exchange of parameters in neighboring replicas during sampling that is helpful in escaping local minima.  Another feature of parallel tempering is their feasibility of implementation in multi-core or parallel computing architectures. In multi-core implementation, factors such as  interprocess communications need to be considered during the exchange  between the neighboring replicas \cite{lamport1986interprocess}, which need to be accounted for when designing parallel tempering for neural networks. 
 
 In the literature, parallel tempering has been used for restricted
 Boltzmann machines \cite{salakhutdinov2007restricted} 
 \cite{hinton2006fast}. Desjardins \emph{et al.} \cite{desjardins2010tempered} showed that parallel tempering is more effective than Gibbs sampling for restricted Boltzmann machines as they lead to faster and better convergence.  Brakel \emph{et  al.} \cite{brakel2012training} extended the method by featuring efficient exchange of information among the replicas and implementing estimation of gradients by averaging over different replicas.  Furthermore, Fischer \emph{et  al.}  \cite{FISCHER2015102} gave an analysis on the   bounds of convergence of parallel tempering for restricted Boltzmann machines. They showed  the significance of geometric spacing of temperature values of the replicas  against linear spacing. 
 
The adoption of Bayesian techniques in estimating neural networks has been slow, because of the challenges in large  datasets and the limitations of MCMC methods for  large scale inference. Parallel tempering overcomes many of these challenges. The within-chain proposals do not necessarily require gradient information, which avoids limitations of gradient-based learning  \cite{pascanu2013difficulty}. Although, random-walk proposals are typically used for parallel tempering, it is worthwhile to explore other proposals such as those based on  Langevin gradients that are used   during  sampling \cite{Chandra2017Langevin}. This approach has shown to greatly enhance MCMC methods for neural networks used for time series prediction. The synergy of Langevin gradients with parallel tempering can alleviate major weaknesses in MCMC methods in terms of efficient proposals required for Bayesian neural learning.  

In this paper, we present a multi-core parallel tempering approach for Bayesian neural networks that takes advantage of high performance computing   for   time series prediction and pattern classification problems. We use Gaussian  likelihood   for prediction and multinomial likelihood for pattern classification problems, receptively. Moreover, we also compare the posterior distributions and the performance in terms of prediction and classification for the selected problems. Furthermore, we investigate the effect on computational time and convergence given the use of Langevin gradients for proposals in parallel tempering. The major contribution of the paper is in the development of parallel tempering for Bayesian neural learning based on parallel computing. 

The rest of the paper is given as follows. Section 2 gives a background on Bayesian neural networks and parallel tempering. Section 3 presents the proposed method while, Section 4 gives design of experiments and results. Section 5 provides a discussion of the results with implications, and section 6 conclusions the paper with directions of future work.

 \section{Background }
 
\subsection{Parallel tempering }

  Parallel tempering (also known as replica exchange or the Multi-Markov Chain method)\cite{swendsen1986replica,hukushima1996exchange,hansmann1997parallel}  has been motivated by thermodynamics of physical systems \cite{swendsen1986replica,patriksson2008temperature} . Overall, in parallel tempering, multiple MCMC chains (known as replicas) are executed at different {\it temperature } values defined by the temperature ladder. The temperature ladder is  used for altering each replica's likelihood function which enables different level of  exploration capabilities. Furthermore, typically the chain at the neighboring replicas are swapped at certain intervals depending on Metropolis-Hastings acceptance criterion.   Typically, gradient free proposals within the replica's are used for proposals for exploring multi-modal and discontinuous posteriors \cite{sen1996bayesian,maraschini2010monte}.   Determining the optimal temperature ladder   for the replicas has been a challenge that attracted some attention in the literature. Rathore \emph{et  al.} \cite{rathore2005optimal} studied the efficiency of parallel tempering in various problems regarding protein simulations and presented an approach for  dynamic allocation of the temperatures. Katzgraber \emph{et  al.}\cite{katzgraber2006feedback} proposed systematic  optimization of 
temperature sets  using an adaptive feedback method that  minimize the round-trip times between the lowest and highest temperatures which effectively increases
efficiency. Bittner \emph{et  al.} \cite{bittner2008make}   showed that by adapting the number of sweeps between replica exchanges, the average round-trip time can be significantly decreased  to achieve  close to 50\% swap rate among the replicas. This increases the efficiency of the parallel tempering algorithm. Furthermore,   Patriksson and Spoel \cite{patriksson2008temperature}  presented an approach    to predict a set of temperatures for use in parallel tempering in the application of molecular biology. 
All these techniques have been applied to specific settings, none of which use a neural network architecture. 

   In the canonical implementation, the exchange  is limited to  neighboring replicas conditioned by the a probability that is determined during sampling. Calvo proposed an alternative technique  \cite{calvo2005all} where the swap probabilities  are calculated a priori and then one swap is proposed. Fielding \emph{et  al.}\cite{fielding2011efficient},  considered replacing the original target posterior distribution with the Gaussian process approximation which requires less computational requirement. The authors replaced true target distribution with the approximation in the high temperature chains while retaining the true target in the lowest temperature chain. Furthermore, Liu \emph{et al.} proposed an approach to reduce the number of replica by  adapting acceptance probability for   exchange  for computational efficiency  \cite{Liu13749}.  

   Although the  approaches discussed  for adapting temperature and improving swapping  are promising, the addition of proposals for methods for temperature values can be computationally expensive. Moreover,  there is no work that has evaluated the respective methods for enhancements on benchmark problems to fully grasp the strengths and weaknesses of the approached.

 
 
 A number of challenges are there when considering multi-core implementations since parallel tempering features exchange or transition between neighboring replicas. One needs to consider efficient strategies that take into account interprocess communication in such systems \cite{LI2009269}.
In order to address this,  Li \emph{et  al.} presented a decentralized  implementation of parallel tempering  was presented that  eliminates global synchronization and
reduces the overhead caused by interprocess communication in exchange of solutions between the chains that run in parallel cores \cite{LI2009269}. Parallel tempering has also been implemented  in a 
distributed volunteer computing network  where computers belonging to the general public are used with help of multi-threading and graphic processing units (GPUs) \cite{karimi2011high}. Furthermore, field programmable gate array (FPGA) implementation of parallel tempering showed much better performance than multi-core and GPU implementations \cite{mingas2017particle}.

 
 Parallel tempering has been used for a number of fields of which some are discussed as follows.  Musiani and Giorgetti  \cite{musiani2017protein}  presented a   review of computational techniques used for simulations of protein aggregation where  parallel tempering was presented as  a widely used technique.  Xie et al. \cite{xie2010parallel} simulated lysozome orientations on charged surfaces using an adaption of the parallel tempering. Tharrington and Jordon \cite{tharrington2003parallel}  used parallel tempering to characterize the finite temperature behavior of $H_{2} O$ clusters.   Littenberg and Neil used parallel tempering for  detection problem in gravitational wave astronomy \cite{littenberg2009bayesian}.  Moreover, Reid et. al used parallel tempering with multi-core implementation  for    inversion  problem for exploration of Earth's resources  \cite{reid2013bayesian}.


 \subsection{Bayesian neural networks }
 
 Bayesian inference provides the  methodology  to update the probability for a hypothesis, called a prior distribution, as more evidence or information becomes available via a likelihood function, to give a posterior distribution.  Bayesian neural networks or neural learning uses the posterior distribution of the weights and biases  \cite{richard1991neural} to make inference regarding these quantities. MCMC techniques are used to get sampling estimates of these  posterior distributions   \cite{mackay1996hyperparameters,robert2014machine}.   
 
 In neural network models, the priors can be informative about the distribution of the weights and biases given the network architecture and expert knowledge. For example it is well known that allowing large values of the weights will put more probability mass on the outcome being either zero or one, \cite{krogh1992simple}. For this reasons the weights are often restricted to lie within the range of [-5,5], which could be implemented using a  uniform prior distribution. Examples of priors informed by prior knowledge include \cite{mackay1995probable,neal2012bayesian,auld2007bayesian}. 
 
The limitations regarding convergence and scalability of MCMC sampling methods has impeded the use of Bayesian methods in neural networks. A number of techniques have been applied to address this issue by incorporating approaches from the optimization literature.  Gradient based methods such as  Hamiltonian MCMC by Neal \emph{et al.} \cite{neal2011mcmc} and Langevin dynamics  \cite{welling2011bayesian},  have significantly improved the rate of convergence of MCMC chains. Chen et al  \cite{chen2016bridging} used   simulated annealing to improve  stochastic gradient MCMC
algorithm for deep neural networks.

In the time series prediction literature,  Liang  \emph{et al.} present an MCMC algorithm for 
neural networks for selected time series problems \cite{liang2005bayesian}, while Chandra \emph{et al.} present  Langevin gradient Bayesian neural networks for prediction \cite{Chandra2017Langevin}.    For  short term time 
series forecasting Bayesian techniques have been used for controlling model 
complexity and selecting inputs in neural networks  \cite{hippert2010evaluation} while Bayesian 
recurrent neural networks \cite{mirikitani2010recursive} have been very effective for time series prediction. Evolutionary algorithms have also been combined with MCMC sampling for 
 Bayesian neural networks   for  time series forecasting 
\cite{kocadaugli2014nonlinear}.

In classification problems,
initial work was done by Wan who provided a Bayesian interpretation for classification with neural networks \cite{wan1990neural}.  
Moving on, a number of successful applications of Bayesian neural networks for classification exist, such as   Internet traffic classification \cite{auld2007bayesian}.

Considering other networks architectures, Hinton \emph{et al.}\cite{hinton2006fast} used complementary priors to derive a fast greedy algorithm  for deep belief networks  to  form an undirected associative memory with application to form a generative model of the joint distribution of handwritten digit images and their labels.   Furthermore, parallel tempering has been used in improving the Gaussian Bernoulli Restricted Boltzmann Machine's (RBMs) train in \cite{cho2011improved}. Prior to this, Cho \emph{et al.} \cite{cho2010parallel} demonstrated the efficiency of Parallel Tempering in RBMs. Desjardins \emph{et al.} utilized parallel tempering for maximum likelihood training of RBMs  \cite{desjardins2010adaptive} and later used it for deep learning using RBMs \cite{desjardins2014deep}. Thus parallel tempering has been vital in development of one of the fundamental building blocks of deep learning - RBMs.  

 \section{Methodology}
 
 In this section, we provide the details for using multi-core parallel tempering for time series prediction and pattern classification problems. The multi-core parallel tempering features two implementations that are different by; 1.)  random-walk proposals , and 2.) Langevin-gradient proposals. We first present  the foundations  followed by the details of the implementations.
 

 \subsection{Model and priors for time series prediction}

Let $y_{t}$ denote a univariate time series. We assume that $y_t$ is generated from a signal plus noise model where the signal is a neural network and the noise is assumed to be i.i.d. Gaussian with variance $\tau^2$, so that,
\begin{equation}
 y_{t} = f({\bf x}_t) + \epsilon_{t}, \; \mbox{for}\;t=1,2,\ldots, n 
 \label{eqn_model1}
 \end{equation}

where $f({\bf x}_t)=E(y_t|{\bf x}_t)$, is an unknown function, ${\bf x}_t = (y_{t-1}, \ldots, y_{t-D})$ is 
a vector of lagged values of $y_{t}$, and $\epsilon_{t}$ is the noise with $\epsilon_t\sim \mathcal N(0,\tau^2)$ $\forall t$.
 \\

We transform into a state-space vector through Taken's theorem 
\cite{Takens1981} which is governed by the embedding dimension (D) and time-lag 
(T). \\
From Taken's Theorem, we define
\begin{equation}
\mathcal{A}_{D,T} = \{t; t > D, \mod(t-(D+1), T) = 0 . \}
\label{takens}
\end{equation}
Let ${\bf y}_{\mathcal{A_{D,T}}}$ to be the set of $y_t$'s for which $t\in\mathcal{A_{D,T}}$, then,   $ \forall  t  \in \mathcal{A_{D,T}}  $.  In this representation the embedding dimension $D$ is equal to the number of inputs in a feed-forward neural network, which we denote by I. 
The expected value of $y_{t}$ given $\bf x_t=(y_1,\ldots,y_{t-1})$  is given by: 
\begin{equation}
f({\bf x}_t)   =   g \bigg(  \delta_o + 
\sum_{h=1}^{H} v_{h} \times g \bigg(  \delta_h + \sum_{i=1}^{I} w_{ih} 
x_{t,i} \bigg)\bigg),
\label{expected_y}
\end{equation}
where $\delta_o$ and $\delta_h$  are the biases for the output and 
hidden $h$ layers, respectively,  $v_h$ is the weight which maps the hidden 
layer $h$ to the output, $w_{ih}$ is the weight which maps $x_{t,i}$ to 
the hidden layer $h$ and $g$ is the activation function, which we assume to be a sigmoid function 
 for the hidden and output layer units of the neural network.
 The parameter vector needed to define the likelihood function contains; the variance of the signal, $\tau^2$, given in Equation \ref{eqn_model1}; the weights of the input to hidden layer $ 
\mathbf {\tilde{w}}$; the weights of the hidden to output layer, $\mathbf {\tilde {v}}$; the bias to the hidden layer ${\boldsymbol \delta_h}$, and the output layer ${\boldsymbol \delta_o}$. There are in total $L = (I*H+H+(O-1)*H+O-1)+1$, parameters, where $I$ is the number of inputs, $H$ is the number of hidden layers, $O$ is the number of classes of the output variable. Note that the number of parameters needed to map a single hidden layer to the output layer is $O-1$.   
\begin{equation}
f_o({\bf x}_t)   =   g \bigg(  \delta_o + 
\sum_{h=1}^{H} v_{ho} \times g \bigg(  \delta_h + \sum_{i=1}^{I} w_{ih} 
y_{t,i} \bigg)\bigg),
\label{expected_y}
\end{equation}
for $o=1,\ldots, O-1$, where $\delta_o$ and $\delta_h$  are the biases for the output $o$ and 
hidden $h$ layer, respectively,  $v_ho$ is the weight which maps the hidden 
layer $h$ to output layer $o$, $w_{ih}$ is the weight which maps $y_{t,i}$ to 
the hidden layer $h$ and $g$ is the activation function, which we assume to be a sigmoid function 
 for the hidden and output layer units of the neural network. The likelihood function is the multivariate normal probability density function and is given by 
\begin{eqnarray}
p({\bf y_{\mathcal{A}_{D,T}}}|\boldsymbol{\theta})&=&-\frac{1}{(2\pi\tau^2)^{n/2}}\times\\
&&\exp\left(
-\frac{1}{2\tau^2}\sum_{t\in\mathcal{A_{D,T}}}\left(y_t-E(y_t|{\bf x}_t)\right)^2\right)\nonumber
\label{loglike}
\end{eqnarray}
where $E(y_t|{\bf x}_t)$ is given by \eqref{expected_y}. 

We  assume that the elements of $\boldsymbol \theta$ are independent {\it apriori}. In addition we assume  apriori that the weights $\mb w$, $\mb v$, and biases $\bs \delta$, have a normal distribution with zero mean and variance $\sigma^2$.  Our prior is now
\begin{eqnarray}
p(\boldsymbol{\theta})& \propto &\frac{1}{(2\pi\sigma^2)^{L/2}}\times\nonumber\\
 &&\exp\Bigg\{-\frac{1}{2\sigma^2}\bigg(\sum_{h=1}^H\sum_{d=1}^D w_{dh}^2+\nonumber\\
  &&\sum_{k=1}^K\sum_{h=1}^H(\delta_{hk}^2+v_{hk}^2)+
  \delta_o^2\bigg) \Bigg\}\nonumber\\
&&\times\tau^{2(1+\nu_1)}\exp\left(\frac{-\nu_2}{\tau^2}\right)
\label{prior_regression}
\end{eqnarray}


\subsection{Model and priors or classification problems}
When the data are discrete, such as in a classification problem, it is inappropriate to model the data as Gaussian.  So for discrete data with $K$ possible classes, we assume that the data, $\mb y=(y_1,\ldots,y_n)$ are generated from a multinomial distribution with parameter vector $\bs \pi=(\pi_1,\ldots,\pi_K)$ where $\sum_{k=1}^K\pi_k=1$. To write the likelihood we introduce a set of indicator variables $z_{i,k}$ where
\begin{equation}
z_{i,k} =
\begin{cases}
    1,& \text{if } y_i = k\\
    0,              & \text{otherwise}
\end{cases}
\end{equation}
for $i=1,\ldots, n$ and $k=1,\ldots,K$. The likelihood function is then  
\begin{eqnarray}
p(\bf y_|\boldsymbol{\pi})&=&\prod_{i=1}^n \prod_{k=1}^K \pi_k^{z_{i,k}}
\label{multinomial}
\end{eqnarray}
for classes $ k = 1,\ldots,K$ where $\pi_k$, the output of the neural network,  is the probability that the data are generated by category $k$. The dependence between this probability and the input features $\mb x$ is modelled as a multinomial logit function 
\begin{equation}
\pi_k = \frac{\exp(f(x_k))}{\sum_{j=1}^K \exp(f(x_j))}
\end{equation}
where $f(x)$ is given by equation\ref{expected_y} and the priors for the weights and biases are given by equation \ref{multinomial}.gm

\subsection{Estimation via Metropolis-Hastings parallel tempering}

In parallel tempering, each replica corresponds to  a predefined  temperature ladder which governs the invariant distribution  where the higher temperature value gives more chance in accepting weaker proposals. 
Hence, parallel tempering \cite{earl2005parallel,hansmann1997parallel} has the feature to sample  multi-modal posterior distributions \cite{neal1996sampling} with temperature ladder where the replicas   can be implemented in distributed or multi-core architectures.   Given $M$ replicas of an ensemble, defined by multiple temperature levels, the state of the ensemble is specified by   $\Theta = ({\theta}_{1},\ldots,{\theta}_{M})$, where $\theta_i$ is the replica at temperature level $T_i$. The samples of $\theta$ from the posterior distribution  are obtained by proposing  values of $\theta^p$, from some known distribution $q(\theta)$.  The chain moves to this proposed value of $\theta^p$ with a probability $\alpha$ or remains at its current location, $\theta^c$, where $\alpha$ is chosen to ensure that the chain is reversible and has stationary distribution, $p(\theta|\mb D)$. 

The development of transitions kernels or proposals which efficiently explore posterior distributions is the subject of much research.   Random-walk proposals feature  a small amount of Gaussian noise to the  current value of the chain $\theta^c$. This has the advantage of having an easier implementation but can be computationally expensive  because many samples are needed for  accurate  exploration.

The Markov chains in the parallel replicas have stationary distributions which are equal to (up to a proportionality constant) $p(\theta|\mb D)^{\beta}$; where $\beta\in[0,1]$, with $\beta=0$ corresponding to a stationary distribution which is uniform, and $\beta=1$ corresponding to a stationary distribution which  is the  posterior.  The replicas with smaller values of $\beta$ are able to explore a larger regions of distribution, while those with higher values of $\beta$ typically explore  local regions. Communication between the parallel replicas is essential for the efficient exploration of the posterior distribution.  This is done by considering the chain and the parameters as part of the space to be explored.  Suppose there are $R$ replicas, indexed by $m$, with corresponding stationary distributions, $p_m(\theta|\mb D)=p(\theta|\mb D)^{\beta_r}$, for, $m=1,\ldots,R$, with  $\beta_1=1$ and $\beta_R<\beta_{R-1}<\ldots,\beta_1$, then the pair $(m,\theta)$ are jointly proposed and accepted/rejected according to the Metropolis-Hastings criterion. The stationary distribution of this sampler is proportional to $p(\theta|\mb D)^{\beta_m}p(m)$. The quantity $p(m)$ must be chosen by the user and is referred to as a {\it pseudoprior}.

\subsection{Langevin gradient-based proposals}

We utilize Langevin-gradients  to update the parameters at each iteration rather than only by using random-walk \cite{BayesianChandraAC17}. The gradients are calculated as follows:

\begin{eqnarray}
\boldsymbol\theta^p &\sim& \mathcal{N}(\bar{\boldsymbol\theta}^{[k]},\Sigma_{\theta}),\;\mbox{where}\;\label{update}\\
\bar{\boldsymbol\theta}^{[k]}&=&\boldsymbol\theta^{[k]} +r\times\nabla E_{\bf y_{\mathcal{A}_{D,T}}}[\boldsymbol\theta^{[k]} ],\label{gradient} \nonumber\\
 E_{\bf y_{\mathcal{A}_{D,T}}}[\boldsymbol\theta^{[k]}]& = & \sum_{t\in{\mathcal{A}_{D,T}}}(y_t-f({\bf x}_t)^{[k]})^2,\nonumber\\
\nabla E_{\bf y_{\mathcal{A}_{D,T}}}[\boldsymbol\theta^{[k]} ]&=& 
\left(\frac{\partial{E}}{\partial{\theta_1}},\ldots, 
\frac{\partial{E}}{\partial{\theta_{L}}}\right)\nonumber
\end{eqnarray}
$r$ is the learning rate, $\Sigma_{\theta}=\sigma^2_{\theta}I_{L}$ and $I_L$ is 
the $L \times L$ identity matrix. So that the newly proposed value of 
$\boldsymbol\theta^p$, consists of 2 parts: \begin{enumerate}
 \item  An gradient descent based weight update given by Equation 
\eqref{gradient}
 \item Add an amount of noise, from $\mathcal{N}(0,\Sigma_{\theta})$.  
 \end{enumerate}

We note that the feature of Langevin-gradients is incorporated in Algorithm \ref{alg:mhptmcmc} where the proposals  consider gradients as given in above equations instead of the random-walk. Note that the Langevin-gradients are applied with a probability (  $L_{prob} = 0.5$  for example). 

\subsection{Algorithm}
This parallel tempering algorithm  for Bayesian neural learning is given in Algorithm~\ref{alg:mhptmcmc}. At first, the algorithm initializes the replicas by following the temperature ladder which is in a geometric progression. Along with this, the hyper-parameters such as the maximum number of samples, number of replicas, swap interval and the type of proposals (random-walk or Langevin-gradients) is chosen. We note that the algorithm uses parallel tempering for global exploration and reverts to canonical MCMC in distributed mode for local exploration. The major change is in the temperature ladder, when in local exploration, all the replicas are set to temperature of 1. The local exploration also  features swap of neighboring replica. Hence, one needs to set the  percentage of samples for global exploration phase beforehand.   The sampling begins by executing each of the replicas in parallel (Step 1). Each replica $\theta$ is updated when the respective proposal is accepted using the Metropolis-Hasting acceptance criterion given by Step 1.3 of  Algorithm~\ref{alg:mhptmcmc}. In case the proposal is accepted, the proposal becomes part of the posterior distribution, otherwise the last accepted sample is added to the posterior distribution as shown in Step 1.3. This procedure is repeated until the replica swap interval is reached ($Swap_{int}$). When all the replicas have sampled till the swap interval,  the algorithm  evaluates if the neighboring replicas need to be swapped by using the Metropolis-Hastings acceptance criterion as done for within replica proposals (Step 2).

\begin{algorithm}[H]
\small
\caption{Metropolis-Hastings Parallel Tempering (MHPT) }\label{alg:mhptmcmc}

\KwResult{Draw samples from $p(\theta|\mb D)$}
1. Set maximum number of samples  ($Max_{samples}$), swap interval ($Swap_{int}$), and number of replicas ($R$) \\
2. Initialize $\theta=\theta^{[0]}$, and $m=m^{[0]}$\;\\
3. Set the current value of $\theta$,  to $\theta^c=\theta^{[0]}$ and $m$ to $m^c=m^{[0]}$\;\\
4. Choose type of proposal (Random-walk (RW) or Langevin-gradients (LG)\\
5. Choose Langevin-gradient frequency (LG-freq)\\ 
6. Set percentage of samples for global exploration phase \\
\While {$Max_{samples}$}
{
 
 \For {$m=1,\ldots,R$   }{
 \For {$k=1,\ldots,$ $Swap_{int}$}{
 \noindent
 
*Local/Global exploration phase \\
\eIf{$global$ is true}{
$R_{temp} = 1/Temp_{m}$\;
}{ 

$R_{temp} = 1$\;

}
 Step 1: Replica sampling \\
 1.1 Propose new sample (solution)\\
 Draw $l\sim U[0,1]$ \\
 \eIf{($LG$ is true) and  ($l < LG-freq$)}{
 Conditional on $m^c$ propose a new value of $\theta$ using Langevin-gradients, 
$\theta^{p}\sim q(\theta|\theta^c,m^c)$, where $q(\theta|\theta^c,m^c)$ is 
given in Equation \ref{update}\;
}{ 
 Conditional on $m^c$ propose a new value of $\theta$, using Random-Walk 
}
1.2 Compute acceptance probability\; 
\[\alpha=\min\left(1,\frac{p(\theta^p|\mb D,m^c)^{\beta_m^c} q(\theta^c|\theta^p,m^c)}{p(\theta^c|\mb D,m^c)^{\beta_m^c} q(\theta^p|\theta^c,m^c)}\right)\]\;

1.3 Acceptance criterion
Draw $u\sim U[0,1]$\;\\ 
\eIf{$u<\alpha$}{
$\theta^{[k]}=\theta^p$\;
}{
$\theta^{[k]}=\theta^c$\;
}
 
}
 Step 2: Replica transition \\
  2.1  Propose a new value of replica swap transition $m^p$, from $q(m^p|m^c)$.\\
2.2 Compute acceptance probability\;

\[\alpha=\min\left(1,\frac{p(\theta^{[k]}|\mb D)^{\beta_{m^p}}p(m^p)q(m^c|m^p)}{p(\theta^{[k]}|\mb D)^{\beta_{m^c}}p(m^c)q(m^p|m^c)}
\right)\]\;
2.3 Neighboring replica swap \\
Draw $u\sim U[0,1]$\;\\
 
\eIf{$u<\alpha$}{
$m^{[k]}=m^p$\;
}{
$m^{[k]}=m^c$\;
}

}
 
}
 \end{algorithm}



\begin{figure*}[htb!]
  \begin{center}
    \includegraphics[width=180mm]{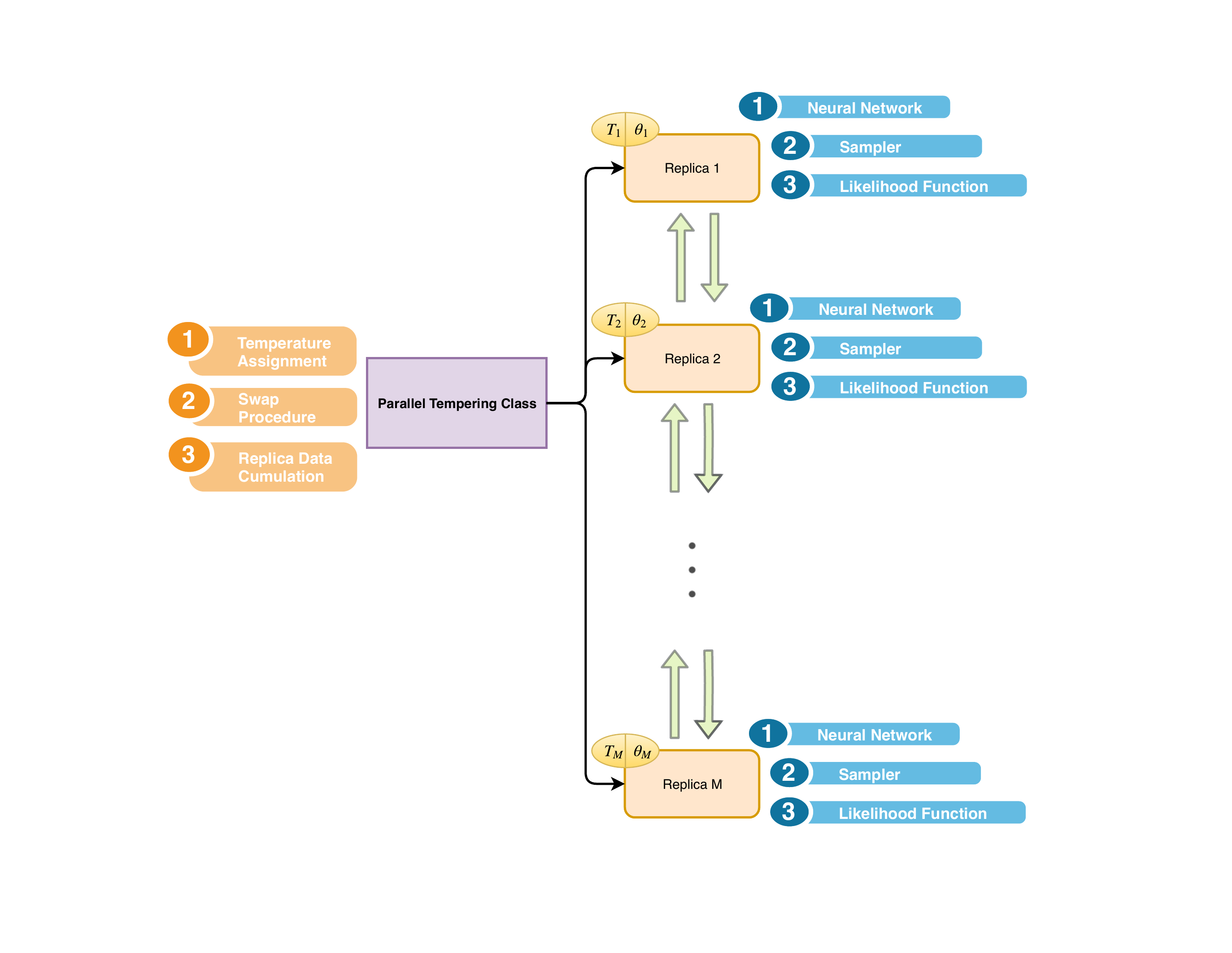} 
    \caption{An overview of  the different replica's that are executed on a multi-processing architecture. Note that the main process controls the given replicas and enables them to exchange the neighboring replicas given the swap time and probability of exchange is satisfied.  }
 \label{fig:pt}
  \end{center}
\end{figure*}

The  multi-core implementation takes into account operating system concepts such as interprocess communications when considering exchange of solutions between the neighboring replicas \cite{lamport1986interprocess}. The implementation is done in  multi-core fashion by which each replica runs on a single  central processing unit (CPU). Figure \ref{fig:pt} gives an overview of  the different replicas that are executed on the multi-processing software wherein each replica runs on a separate core with inter-process communication for exchanging neighboring replicas. The main process runs on a separate core which  controls the  replicas and enables them to exchange the neighboring replicas given the swap interval and probability of exchange is satisfied. Each replica  is allocated a fixed sampling time controlled by the number of iterations. The main process waits for all samplers to complete their required sampling till the swap interval iteration after which the samplers attempt configuration exchange. Once the replica reaches this junction, the main process   proposes configuration swaps between adjacent samplers based on likelihoods of adjacent chains. Main process notifies the replicas post swapping to resume sampling with latest configurations in the chain for each replica. The process continues with the sampling and proposing swaps until the the maximum sampling time.

We utilize  multi-processing software development packages for the resources to have efficient inter-process communication \cite{lamport1986interprocess} taking into account the swapping of replicas at certain intervals in the iterations. We implement our own parallel tempering for multi-core architecture using the Python multiprocessing library \cite{singh2013parallel}  and release an open-source  software package for the same given here \footnote{ Langevin-gradient parallel tempering for Bayesian neural learning: \href{https://github.com/sydney-machine-learning/parallel-tempering-neural-net}{https://github.com/sydney-machine-learning/parallel-tempering-neural-net}}. The software we developed is very generic and can be modified easily to suit a large set of applications.

\section{Experiments and Results}

This section presents the experimental evaluation of multi-core parallel tempering  for Bayesian neural learning given selected time series prediction and classification problems.

\subsection{Experimental Design}

We present experimental  design that features  parallel tempering based on random-walk (PT-RW)  and  Langevin gradient (PT-LG) proposals based on \cite{Chandra2017Langevin}. We use geometric temperature spacing \cite{kone2005selection} to determine the temperature for each replica of the respective algorithms. In all experiments, we use a hybrid parallel tempering implementation where parallel tempering is used in the first stage that compromises of  60 percent of the sampling time. In the second stage, the framework changes to canonical MCMC where the temperature value of the replicas become 1.  In this way, parallel tempering is used mostly for exploration or global search  in the first phase which is  followed by exploitation or local search  in the second phase.
Moreover, first 50 percent of samples are discarded as burn-in period required for convergence. This is a standard procedure in sampling methods used for Bayesian inference. Furthermore, the Langevin-gradient proposals are applied with a probability of 0.5 for every sample proposed. The effect of the learning rate used for the Langevin-gradient are evaluated in the experiments. The experiments are designed as follows. 

\begin{itemize}
\item Evaluate effect of maximum temperature for a selected time series problem;
\item Evaluate effect of swap interval for a selected time series problem; 
\item Evaluate effect of Langevin-gradient rate  for a selected time series problem; 
\item Compare PT-RW and PT-LG  for a range  time series prediction problems; 
\item  Compare PT-RW and PT-LG for  pattern classification problems.
\end{itemize}


We used  one hidden layer feedforward neural network where the number of hidden units was $h=5$ for time series prediction problems. In the case of pattern classification, the number of hidden layers is provided in Table \ref{tab:datas}.  We evaluate the effect of the  hyper-parameters  in parallel tempering, which include the maximum geometric  temperature spacing  and swap interval for swapping neighboring replicas. Note that all experiments used 10 replicas with a multi-core implementation that was experimented using high-performance computing environment in order to ensure that each replica is executed on a separate core. 

In random-walk proposals, we draw  and add a Gaussian noise to the  weights and biases of the network  from a standard normal distribution, with mean of 0 and standard deviation of  $0.025$. As given in Equation \ref{expected_y} whereas for $\eta$ (where $\eta$ is defined as $\eta = log(var(y-E(y|{\bf x}))$), the random-walk step is $0.2$. The random-walk step becomes the standard deviation when we draw the weights and $\eta$ from the normal distribution centered around $0$. The parameters of the priors (see Equation \eqref{prior_regression}) were set as ${\sigma}^{2} = 25, {\nu}_{1} = 0$ and ${\nu}_{2} = 0$. 

\subsubsection{Time series problems}

The benchmark problems used are Sunspot, Lazer, Mackey-Glass, Lorenz, Henon and Rossler time series. Takens' embedding theorem \cite{Takens1981} is applied with selected values as follows. Dimension, $D = 4$ and time lag, $T = 2$ is used for reconstruction of the respective time series into state-space vector as done in previous works \cite{Chandra2018_CMTL,chandra2017_CMTL}. All the problems used first one thousand data points of the time series from which 60\% was used for training and remaining for testing. The prediction performance is  measured by root mean squared error as follows

$$RMSE = \sqrt{\frac{1}{N} \sum_{i=1}^{N} (y_i - \hat{y_i})^2}$$

where $y_i$ and $\hat{y_i}$ are the true value and the predicted value respectively. $N$ is the total length of the data. The maximum sampling time is set to 100,000 samples for the respective time series problems.

\subsubsection{Pattern classification problems}

We selected six pattern classification problems from University of California Irvine machine learning repository \cite{AsuncionNewman2007}. The problems were selected according to their size and complexity in terms of number of features and samples. We experiment with small datasets with small number of features (Iris and  Cancer), then with small dataset with high number of features (Ionosphere). Moreover,  three relatively large datasets with high number features, one with low number of classes (Bank-Market) and others with high number of classes (Pen-Digit and Chess) as shown in Table \ref{tab:datas}. 

\begin{table}[h]
\centering
\small
 \caption{Classification Dataset Description}
\label{tab:datas}
\begin{tabular}{lllll}
\hline
 \hline
Dataset&Instances&Attributes&Classes&Hidden Units\\
\hline
\hline
Iris&150&4&3&12\\
Breast Cancer&569&9&2&12\\
Ionosphere&351&34&2&50\\
Bank-Market&11162&20&2&50\\
Pen-Digit&10992&16&10&30 \\  
Chess& 28056 & 6& 18& 25 \\

\hline
\hline
\end{tabular}
\end{table}

\subsection{Results: time series prediction}

\begin{table*}[h]
\centering
\small
 \caption{Evaluate maximum geometric temperature}
\label{tab:result_temp}
\begin{tabular}{llllll}
\hline
 \hline
 Max. temp.&Train  (mean, best, std) &Test  (mean, best, std)    & Swap per. & Accept per. &  Time (min.) \\
\hline
\hline  

  2  &   0.0639 0.0208 0.0292& 0.0543 0.0184 0.0244 &31.97  &30.26 & 3.65  \\
 4 &   0.0599 0.0206 0.0299 &0.0523 0.0176 0.0248& 38.21  &33.74  &3.94  \\
 6 &  0.0656 0.0220 0.0299 &0.0576 0.0210 0.0277 &48.59 & 36.01 & 4.70 \\
  8 & 0.0692 0.0277 0.0311& 0.0584 0.0209 0.0261& 44.97  &37.72  &3.79  \\ 
  10 &   0.0635 0.0250 0.0293 &0.0547 0.0212 0.0251& 43.97 & 38.14 & 3.73 \\

\hline
\end{tabular}
\end{table*}

\begin{table*}[h]
\centering
\small
 \caption{Evaluate swap interval}
\label{tab:result_swap}
\begin{tabular}{llllll}
\hline
 \hline
Swap interval &Train  (mean, best, std) &Test  (mean, best, std)    & Swap per. & Accept per. &  Time (min.) \\
\hline
\hline

  100&  0.0628 0.0233 0.0286 & 0.0554 0.0213 0.0264 & 31.33   & 34.19  &3.66  \\
200 & 0.0608 0.0226 0.0308 & 0.0521 0.0194 0.0262 & 40.65  & 33.41  &3.49 \\
 300 & 0.0657 0.0190 0.0328 &0.0562 0.0173 0.0263 &39.01  &34.38  & 3.47 \\ 
 400 &  0.0648 0.0238 0.0281 &0.0552 0.0214 0.0230& 40.11 & 34.35 &3.68 \\
  500 &   0.0676 0.0220 0.0310 &0.0577 0.0190 0.0279 &50.00  &34.82 & 3.83  \\
  600 &0.0621 0.0255 0.0266 &0.0549 0.0209 0.0247 &50.00   &32.62 & 3.54  \\
 700 & 0.0614 0.0197 0.0320& 0.0522 0.0169 0.0280 &47.36 & 34.10& 3.52 \\
  800 &  0.0739 0.0217 0.0291 &0.0666 0.0214 0.0264 &31.57  &36.08 & 3.52 \\

\hline
\end{tabular}
\end{table*}

\begin{table*}[h]
\centering
\small
 \caption{Evaluate LG Rate}
\label{tab:result_lgrate}
\begin{tabular}{llllll}
\hline
 \hline
LG-frequency &Train  (mean, best, std) &Test  (mean, best, std)    & Swap per. & Accept per. &  Time (min.) \\
\hline
\hline  
   0&  0.0628 0.0233 0.0286 & 0.0554 0.0213 0.0264 & 31.33   & 34.19  &3.66  \\
 0.1 & 0.0539 0.0224 0.0286 & 0.0500 0.0238 0.0273& 34.20 & 30.50  &6.06  \\
 0.2 & 0.0361 0.0099 0.0261 & 0.0331 0.0083 0.0239 & 45.66  &26.33  &7.33 \\
  0.3  & 0.0346 0.0096 0.0226 &0.0318 0.0077 0.0233& 39.42  &23.84  &9.10 \\
 0.4  & 0.0365 0.0113 0.0249 &0.0340 0.0107 0.0232 &45.57  &22.74  &10.12  \\
  0.5 &   0.0451 0.0188 0.0266& 0.0402 0.0177 0.0248& 32.88  &21.19 & 11.6  \\ 
 0.6  & 0.0464 0.0247 0.0245& 0.0433 0.0205 0.0234& 52.32  &19.65  &13.22  \\
 0.7  & 0.0396 0.0232 0.0234& 0.0373 0.0214 0.0202 &48.65  &18.74 & 15.23 \\ 
 0.8 &  0.0350 0.0195 0.0230 &0.0346 0.0168 0.0231 &45.13  &16.66 & 17.63  \\

\hline
\end{tabular}
\end{table*}


 We first show results for effect of different parameter values for selected time series problem. Table \ref{tab:result_temp} shows the performance of the PT-RW method for Lazer problem given different values of the maximum temperature  of the geometric temperature ladder given fixed swap interval of 100 samples. We note that the maximum temperature has a direct effect on the swap probability. Higher values would implies  that some of the replicas with high values of the temperature gives more opportunities  for exploration as it allows the replica to get out of the local minimum while the replicas with lower temperature values focus on exploitation of a local region in the likelihood landscape. The results show that temperature value of 4 gives the best results  in prediction denoted by  lowest RMSE for training and test performance   considering both the train and test datasets. It seems that the problems requires less lower swap rate as it needs to concentrate more on exploiting local region. Moving on, Table \ref{tab:result_swap} shows the results for the changes in the swap interval that determines how often to calculate the swap probability for swapping given maximum geometric temperature of 4. Note that during swapping, the master process freezes all the replica sampling process and  calculates the swap probability in order to swap between neighboring replicas. We note that there is not a major difference in the performance given the range of swap interval considered.  Howsoever, the results show that the swap interval of 200 is best for time series problems. Furthermore, Table \ref{tab:result_lgrate} shows the effects of the Langevin-gradient frequency  (LG-frequency) where swap interval of 100 samples was used with  learning rate of 0.1. We notice that the higher rate of using gradients review more time. This is due to the computational cost of calculating gradients for the given proposals. The LG-frequency of [0.2 - 0.4] shows   the best results in prediction with moderate computational load. These values may be slightly different for the different problems.


Considering  the hyper-parameters (maximum temperature, swap interval and LG-frequency) from previous results, Table \ref{tab:result_timeseries} provides results for PT-RW and PT-LG for the selected problems.  Note that we provide experimentation of two experimental setting for  PT-LG (learning rate) where learning rate of 0.1 and 0.01 are used. The results   show that PT-LG (0.1) gives the best performance in terms of training and test performance (RMSE) for Lazer, Sunspot, Lorenz and Henon problems. In other problems (Mackey, Rossler), it achieves similar performance when compared to PT-RW and PT-LG (0.01).  We note that a highly significant improvement in results is made for the Henon and Lazer problems with PT-LG (0.1).  We can  infer  that the learning rate is an  important parameter    to harness the advantage of Langevin-gradients. In terms of computational time, we notice that PT-LG is far more time consuming than PT-RW, due to the cost of calculating the gradients. 

In terms of percentage of accepted proposals over the entire sampling process,  PT-LG (0.1) has least  acceptance percentage of the proposals. On the other hand, we notice that PT-RW has the highest acceptance percentage followed by  PT-LG (0.01) . Moreover, the swap percentage is generally higher for  PT-LG (0.1) when compared to others for most of the problems. It could be argued that the higher rate of  accepted proposals deteriorates the results. This could imply that weaker proposals are accepted more often in these settings, hence there is emphasis on exploration rather than exploitation. The higher swap percentage indicates exploration over local minimums.  Furthermore,  small learning rate  (0.01)  forces the gradient to have little influence on the proposals and hence  PT-LG (0.01) behaves more like  PT-RW.

\begin{table*}[h]
\centering
\small
 \caption{Prediction Results}
\label{tab:result_timeseries}
\begin{tabular}{lllllll}
\hline
 \hline
Dataset&Method&Train  (mean, best, std) &Test  (mean, best, std)    & Swap per. & Accept per. &  Time (min.) \\
\hline
\hline  
 
Lazer&PT-RW& 0.0640 0.0218 0.0325 &0.0565 0.0209 0.0270& 42.26& 35.31 &4.53 \\   
  &PT-LG (0.10)&   0.0383 0.0187 0.0240 & 0.0353 0.0161 0.0212 & 48.45 &  19.91 & 11.53\\ 
   &PT-LG (0.01)&  0.0446 0.0160 0.0283 &0.0414 0.0160 0.0253& 51.45 & 30.97 & 11.50 \\
\hline 

 Sunspot &PT-RW&     0.0242 0.0041 0.0170 & 0.0239 0.0050 0.0161 & 44.45 &  18.30&  4.82\\
  
 &PT-LG(0.10) & 0.0199 0.0031 0.0155 & 0.0192 0.0033 0.0146 &  48.45&  12.57 &  11.61 \\
 
   &PT-LG (0.01)& 0.0215 0.0032 0.0168& 0.0204 0.0034 0.0154 &46.94 &15.16 & 11.47 \\
 
\hline    
 Mackey &PT-RW &0.0060 0.0005 0.0051 & 0.0061 0.0005 0.0051 &  42.11& 8.19& 4.59   \\  
 &PT-LG (0.1)& 0.0061 0.0009 0.0047 & 0.0062 0.0009 0.0048 &  49.10 & 5.72& 11.68   
 \\ 
 &PT-LG (0.01)&  0.0064 0.0008 0.0052 & 0.0065 0.0008 0.0053 & 48.58 & 8.38 & 11.43    \\
 
\hline  
 Lorenz &PT-RW& 0.0192 0.0033 0.0113& 0.0171 0.0037 0.0094& 39.48 &14.48& 4.45 \\  
 &PT-LG (0.1) &0.0181 0.0018 0.0117 &0.0157 0.0018 0.0094 &50.37 & 9.66&  11.48   \\ 
  & PT-LG (0.01) & 0.0173 0.0023 0.0123 & 0.0147 0.0024 0.0095&  46.30 & 11.91 & 11.42   \\

\hline 
 Rossler&PT-RW & 0.0173 0.0011 0.0144 & 0.0175 0.0011 0.0148 &48.11 & 12.53 & 4.22 \\  
 &PT-LG (0.1) & 0.0172 0.0008 0.0154 &0.0175 0.0009 0.0155 &39.57 &  8.58 & 11.60 \\ 
 &PT-LG (0.01) &  0.0171 0.0011 0.0136 & 0.0173 0.0011 0.0135 &  50.18 &  10.98 & 11.36  \\
 
\hline 
 Henon &PT-RW  &   0.1230 0.0296 0.0167& 0.1198 0.0299 0.0161& 48.58 &38.08 & 4.21 \\  
 &PT-LG (0.1) & 0.0201 0.0025 0.0146 &0.0190 0.0029 0.0131 &47.43& 11.39 & 11.41  \\
 &PT-LG (0.01) & 0.0992 0.0347 0.0221 & 0.0963 0.0341 0.0209 & 36.04 &  36.27 & 11.49  \\

\hline
\end{tabular}
\end{table*}

\subsection{Results: classification}

The multinomial likelihood given in Equation \ref{multinomial} is used for pattern classification experiments that compare PT-RW and PT-LG. We used maximum sampling time of 50,000 samples with LG-frequency  of 0.5, swap interval of 100 and maximum temperature of 10.  The results for the classification performance   along with the time taken and acceptance for the respective  dataset  is provided in Table \ref{tab:result_class}.  We use PT-LG(learning rate) of 0.01 and 0.1 and compare the performance with random-walk (PT-RW) proposals. The results show that PT-LG(0.01) overall achieves the best training and generalization performance for all the given problems.  PT-LG(0.1) has similar training and test performance when compared to PT-RW for  certain problems (Bank and Pen-Digit) only. In majority of the cases PT-LG(0.1) outperforms PT-RW.  We notice that the smaller datasets (Iris, Ionosphere and Cancer) have  much higher acceptance percentage of proposals when compared to larger datasets. Furthermore, the higher computational cost of applying Langevin-gradients is evident in all the problems.

\begin{table*}[h]
\centering
\small
 \caption{Classification Results}
\label{tab:result_class}
\begin{tabular}{llllllll}
\hline
 \hline
Dataset&Method&Train  (mean, best, std) &Test  (mean, best, std)     & Swap perc. &Accept perc. &Time (min.) \\
\hline
\hline 
 
 Iris&PT-RW&  51.39 15.02 91.43 & 50.18 41.78 100.00 &52.56 &95.32 &1.26\\ 
&PT-LG ( 0.1) &64.93 21.51 100.00 &59.33 39.84 100.00 &51.08 &51.48& 1.81\\
 &PT-LG (0.01) &97.32 0.92 99.05 &96.76 0.96 99.10 & 51.77 &97.55& 2.09\\
  
\hline 

Ionosphere&PT-RW& 68.92 16.53 91.84& 51.29 30.73 91.74& 50.61 &89.32& 3.50\\  
&PT-LG (0.1)&  65.78 10.87 85.71& 84.63 9.54 96.33& 47.83 &45.46 &4.70 \\
 &PT-LG (0.01) &  98.55 0.55 99.59 &92.19 2.92 98.17& 51.77& 92.40 &5.07\\  
\hline  
Cancer&PT-RW&83.78 20.79 97.14 &83.55 27.85 99.52& 40.18 &89.71 &2.78 \\
&PT-LG (0.1)& 83.87 17.33 97.55 & 90.59 16.67 99.52 & 41.71 &  43.87 & 5.13	 \\ 
 &PT-LG(0.01)& 97.00 0.29 97.75 &98.77 0.32 99.52& 49.25& 94.67 &5.09  \\
  
\hline 
 
 Bank &PT-RW&  78.39 1.34 80.11& 77.49 0.90 79.45 &49.13 &61.59& 27.71\\ 
&PT-LG(0.1) &  77.90 1.92 79.71 & 77.74 1.27 79.57 & 46.17 &29.61 &69.89  \\	 
 &PT-LG(0.01) &   80.75 1.45 85.41 &79.96   0.81 82.61& 50.00 & 31.50 & 86.94\\

\hline 

 Pen-Digit &PT-RW& 76.67 17.44 95.24& 71.93 16.59 90.62 &45.60 &50.72 &57.13 \\  
 &PT-LG(0.1) &  73.91 17.36 91.98 &70.09 16.08 85.68 &46.89& 24.95 &87.06  \\	
 &PT-LG(0.01) &   84.98 7.42 96.02 & 81.24 6.82 91.25 & 51.08 &25.09& 86.62\\ 
\hline

 Chess &PT-RW  &   89.48 17.46 100.00 &90.06 15.93 100.00 &48.09& 69.09& 252.56 \\ 
 &PT-LG(0.1) &   100.00 0.00 100.00 & 100.00 0.00 100.00 & 49.88 & 92.61 &327.45  \\ 
 &PT-LG(0.1) & 100.00 0.00 100.00 & 100.00 0.00 100.00 &50.12 &88.87 & 323.10\\

\hline
\end{tabular}
\end{table*}

\begin{figure*}[h]  
\begin{subfigure}
\centering
\includegraphics[width=70mm]{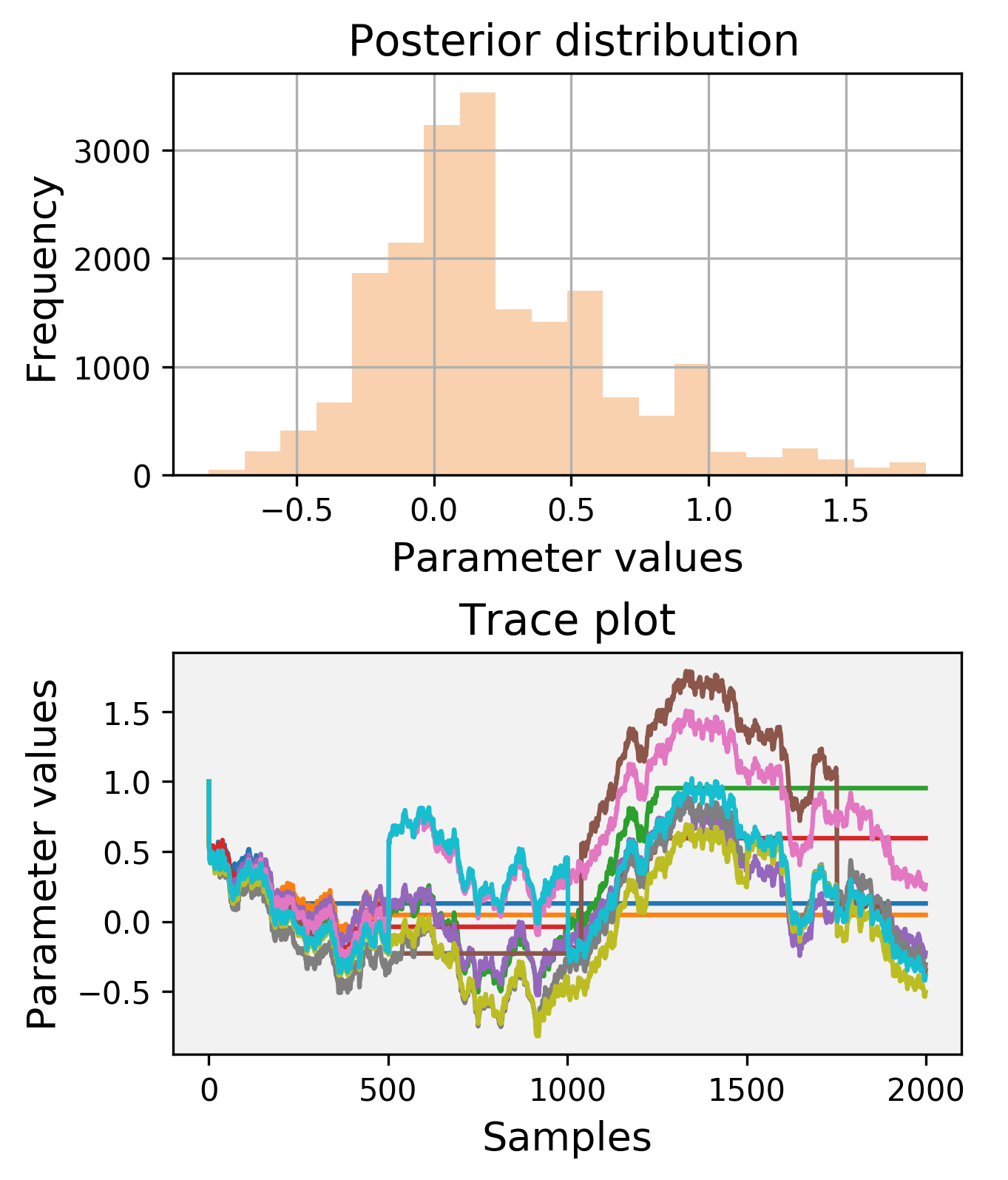}     
\end{subfigure}
\begin{subfigure}
\centering
\includegraphics[width=70mm]{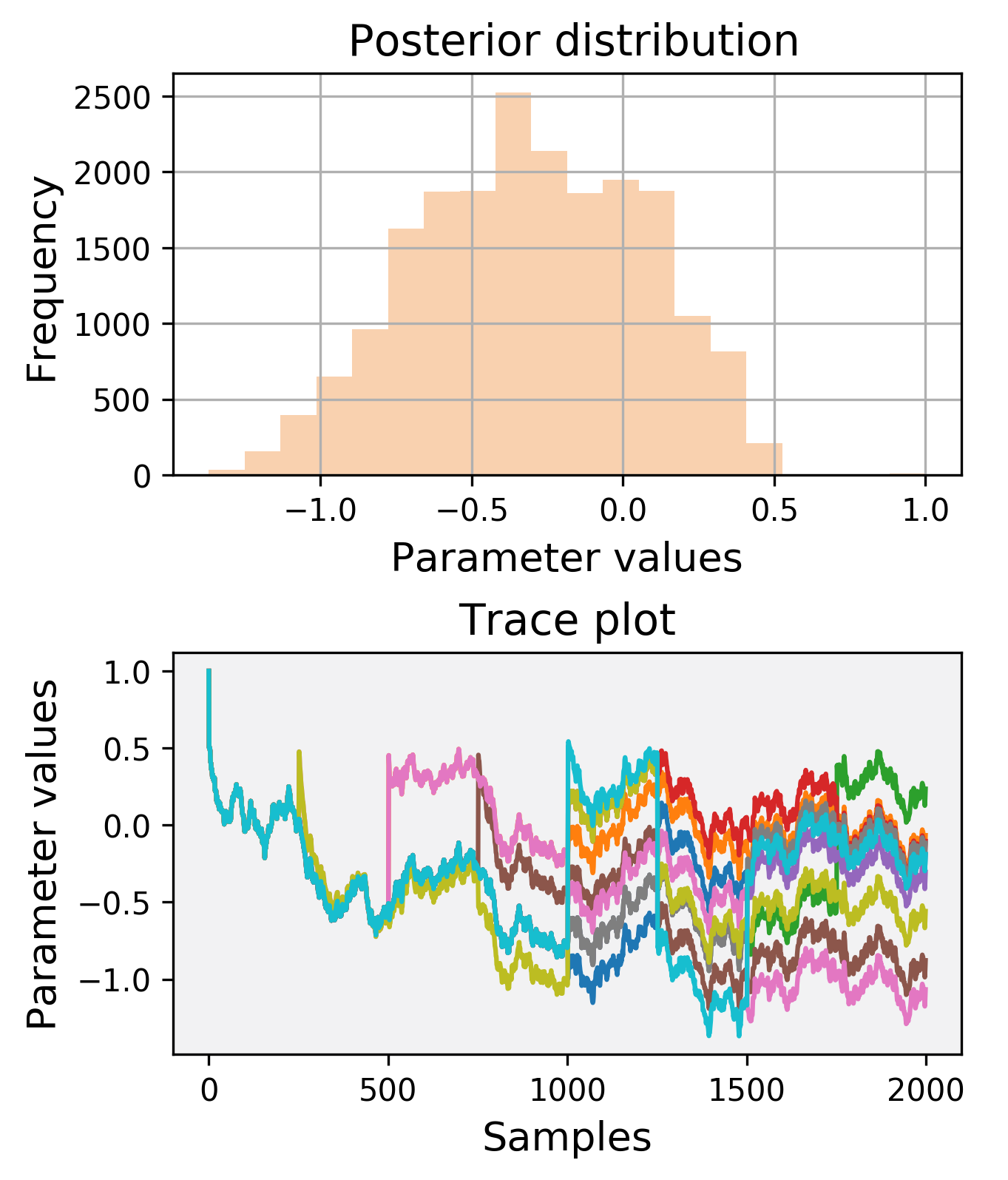}     
\end{subfigure}
\centering
\caption{Posterior distribution with sampling  trace-plot   of a selected  weight  in the input-hidden layer for Iris problem given by PT-LG. }
\label{fig:posterior16}
\end{figure*}

\section{Discussion} 
 We presented a multi-core implementation of parallel tempering with variations such as random-walk and Langevin-gradients for time series prediction and pattern classification problems. The goal is to harness the advantage of   parallel computing for large datasets where  thousands of samples are required for convergence for a posterior distribution. We demonstrated the concept using smaller time series prediction datasets in order to effectively  study the effects of certain hyper-parameters.  The experiments revealed that it is important to have the right hyper-parameter setting such as swap-interval and maximum temperature for geometric temperature spacing. Moreover, the learning rate used for Langevin-gradients can also be an issue given the type of problem, which also depends on the nature of the time series; whether it is synthetic or real-world problem and how much of  noise and inconsistencies are present. We noticed that the Langevin-gradients has more effect since it improved the results further for real-world time series (Lazer and Sunspot problems). In certain cases, higher level of noise and chaotic behavior is present in synthetic time series, such as the Henon time series. In this case, the Langevin-gradient significantly improved performance. One needs to take into account the computational costs of calculating the gradients as shown in the results. Earlier, we found that rate of [0.2 - 0.3] gives the best performance  in terms of accuracy in prediction and also computation time for time series problems. Hence, it is worthwhile   to use lower  values of LG-frequency.   Furthermore, our implementation and experiments were limited to ten replicas, which could be increased for bigger datasets and problems.

 In the case of classification problems, we observe that both variations of Langevin-gradient parallel tempering  significantly improved the classification performance for majority of the problems. In some cases, the classification performance was slightly improved, however, Langevin-gradients have used high computational time in general. In general, we noticed that lower learning rate (0.01) had best performance for classification problem, but had little effect for the time series prediction problems. Therefore, the sampling is highly sensitive to the Langevin-gradient learning rate which needs to be tailored for the type of the problem. Furthermore, note that the rate at which the Langevin-gradient is applied adds to the computational cost. We need to further highlight that the parallel tempering is used mostly for exploration in the first phase of sampling and most of these values are discarded given  50 percent of \textit{burnin} period. The second phase follows canonical MCMC sampling, while taking advantage of swap and also multi-core implementation for lowering computational time. 
 
 We tested small scale to medium scale classification problems that considered a few hundred to thousands of weights in the neural networks. Large number of neural network weights implies large scale inference which has been a challenge of Bayesian neural networks that have mainly been limited to small scale problems. Multi-core parallel tempering has demonstrated to have potential to be applied to deep learning that involve recurrent neural networks and convolution networks that typical deal with thousands to millions of weights. Although, millions of weights is a huge task for vision related problems, given the incorporation of Langevin-gradients, this could be possible. Another possibility is to use stochastic gradient descent implementation for Langevin-gradients with parallel tempering. We only considered calculation of gradient for entire datasets  which could be difficult for big data problems.  The deep learning literature has been constantly trying to improve learning methods that have novel methodologies to feature gradient information which can be used to give better proposals for inference via parallel tempering. This further opens up Bayesian inference for new neural network architectures such as long short-term memory recurrent networks (LSTMs) and  generative adversarial networks   \cite{hochreiter1997long,goodfellow2014generative}. Furthermore, the use  of efficient priors  via transfer learning methods could also be a direction, where the priors feature knowledge learn from similar problems. Although uncertainty quantification is not a major requirement in  some of the computer vision problems, in specific domains such as medical imaging and security, it is important to feature a Bayesian perspective that can be used to fuse many aspects of the data. For instance, three-dimensional (3D) face recognition could provide data from different sensors and cameras that can be fused given a Bayesian methodology. Furthermore, the framework proposed could be used for other models, such as in Earth science, which have expensive models that require efficient optimization or parameter estimation techniques \cite{chandra2018bayeslands}. Related work has explored the use of such framework for estimation of Solid Earth evolution models. In such problems, due to the complexity of the models, there is no scope for incorporating gradients since they are unavailable. In such cases, the need to develop heuristic methods for acquiring gradients is important. Hence, meta-heuristic and evolutionary algorithms could be incorporated with parallel tempering in models that do not have gradients. This could also apply to dynamic and complex neural network architectures, which has in past been addressed through neuro-evolution \cite{Potter_Jong2000,chandra2017_CMTL}.

\section{Conclusions and Future Work}

The major contributions of the paper can be highlighted as follows. 1.) the application of multi-core parallel tempering for enhancing Bayesian neural networks; 2.) the use of Langevin-gradient based proposals to enhance proposals in parallel tempering; 3.) and the application of the methodology to prediction and classification problems of varying degrees of complexity with performance evaluation in terms of the quality of decision making. The experimental analysis have shown that  Langevin-gradient parallel tempering significantly improves the convergence with better prediction and classification performance. Moreover, additional computational costs are needed for Langevin-gradients and in future work such challenges can be tacked with more efficient gradient proposals. This motivates the use of the methodology for large scale problems that faced challenges with canonical implementations of Bayesian neural networks. 

In future work, the implementation of parallel tempering in multi-core architectures can be applied to deep neutral network architectures and various applications that require uncertainty quantification in prediction or decision making process.

\section*{Acknowledgements}

We would like to thanks Artemis high performance computing support at University of Sydney and Arpit Kapoor for providing technical support.

\bibliographystyle{model1-num-names}

\bibliography{feature.bib,rr,aicrg,cyclone,2018,Chandra-Rohitash,Bays,sample}

\end{document}